\documentclass[runningheads]{llncs}

\usepackage{hyperref}
\usepackage{graphicx}
\usepackage{amsmath}
\usepackage{amsthm}
\usepackage{booktabs}
\usepackage{algorithm}
\usepackage{algorithmic}
\usepackage{amssymb}
\usepackage{color}
\usepackage{multirow}
\usepackage{latexsym}
\usepackage{bm}
\usepackage{subfigure}
\usepackage[dvipsnames]{xcolor}
\usepackage{colortbl}
\usepackage{balance}
\usepackage{ulem}
\usepackage{setspace}
\usepackage{lipsum}

\newcommand{\chenzhuo}[1]{{\color{black}#1}}

\newcommand{\zw}[1]{{\color{black}#1}}

\begin{document}
\title{Target-oriented Sentiment Classification with Sequential Cross-modal Semantic Graph}
\author{
Yufeng Huang\inst{1}
\and
Zhuo Chen\inst{2}
\and
Jiaoyan Chen\inst{3}
\and
Jeff Z. Pan\inst{4}
\and
Zhen Yao\inst{1}
\and
Wen Zhang\inst{1}\thanks{Corresponding author.} 
}
\institute{School of Software Technology, Zhejiang University
\and
College of Computer Science and Technology, Zhejiang University
\and
Department of Computer Science, The University of Manchester\\
\and
School of Informatics, The University of Edinburgh
\email{\{huangyufeng,zhuo.chen,yz0204,zhang.wen\}@zju.edu.cn}
\email{jiaoyan.chen@manchester.ac.uk, j.z.pan@ed.ac.uk}
}
\authorrunning{Y. Huang et al.}
\maketitle             
%
\begin{abstract}
Multi-modal aspect-based sentiment classification (MABSC) is 
task of classifying the sentiment of a target entity mentioned in a sentence and an image. However, previous methods failed to account for the fine-grained semantic association between the image and the text, which resulted in limited identification of fine-grained image aspects and opinions. To address these limitations, in this paper we propose a new approach called SeqCSG,  
which enhances the encoder-decoder sentiment classification framework using sequential cross-modal semantic graphs. SeqCSG utilizes image captions and scene graphs to extract both global and local fine-grained image information and considers them as elements of the cross-modal semantic graph along with tokens from tweets. The sequential cross-modal semantic graph is represented as a sequence with a multi-modal adjacency matrix indicating relationships between elements. Experimental results   show that the approach outperforms existing methods and achieves state-of-the-art performance on two standard datasets. Further analysis has demonstrated that the model can implicitly learn the correlation between fine-grained information of the image and the text with the given target. 
Our code is available at \url{https://github.com/zjukg/SeqCSG}.

\end{abstract}


\section{Introduction}
Multi-modal aspect-based sentiment classification (MABSC) is an emerging task 
of classifying the sentiment of a given target such as a mentioned entity in data with different modalities. Specifically, MABSC seeks to identify the sentiment polarities of a target when given a text-image pair.

\begin{figure}[t]
  \centering
  \includegraphics[width = 0.95\linewidth]{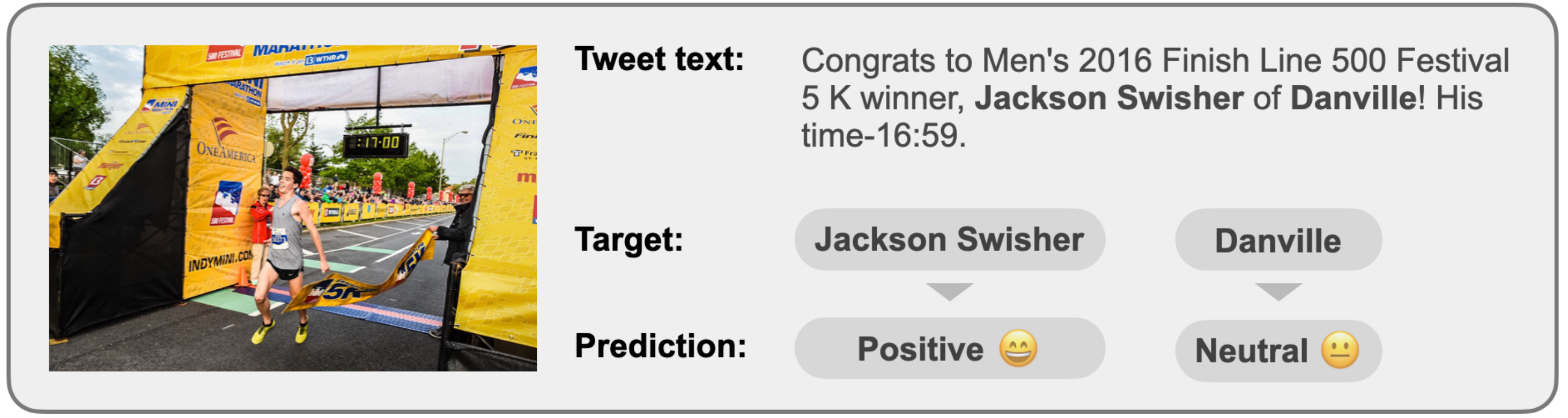}
  \vspace{-5pt}
  \caption{Examples of MABSC. 
  }
  \label{fig:intro}
  \vspace{-10pt}
\end{figure}


Recent years have witnessed increasing attention on the MABSC task and many methods are proposed for this challenging task. 
\zw{Some studies \cite{DBLP:conf/mm/0001F21}}
fuse caption and tweet to achieve model alignment. Yu et al. \cite{DBLP:conf/ijcai/YuWXL22} proposes a multi-task learning model to capture the image-target matching relations. Zhao et al.\cite{DBLP:conf/coling/ZhaoWLDHC22} leverages the adjective-noun pairs to align text and image. Other works like \cite{DBLP:conf/acl/LingYX22} model aspects, opinions and their alignment through task-specific visual language pre-training (VLP-MABSA). 
\zw{These methods mainly relied on coarse-grained information extracted from images, such as the features of the entire image, and achieved alignment between images and texts to a certain extent. However, it is very common to have the same image and text but different targets in the MABSC task. While coarse-grained features are} insufficient for accurately classifying two tasks with the same image-text pair but different targets and sentiments, as shown in Figure \ref{fig:intro}.
Therefore, it is crucial to model both the global and local fine-grained information from the image, while also leveraging text that takes into account the target and the fine-grained information from the image in a cross-modal manner.

With the objective of achieving this aim, we extract global and local features as fine-grained image information by utilizing image captions and scene graphs, respectively. We then propose a method to construct a sequential cross-modal semantic graph for each image-text pair, which is represented as a sequence with a multi-modal adjacency matrix. This representation enables us to obtain a high-level structured representation of the visual context.
Specifically, the elements of the sequential cross-modal semantic graph include tokens of the input text and the image caption, as well as triples that indicate relationships between fine-grained images and objects in the scene graph.
Then we transform all these elements into a sequence and construct the structure of the semantic graph through a multi-modal adjacency matrix indicating the connections between different elements. Meanwhile, we built a manual prompt template that guides the model to connect the target and the other information.
To make effective use of the sequential cross-modal semantic graph, we introduce an encoder-decoder framework that incorporates a target prompt template.

To demonstrate the effectiveness of our approach, we experimentally evaluate the model on two benchmarks, Twitter2015 \cite{DBLP:conf/aaai/0001FLH18} and Twitter2017 \cite{DBLP:conf/acl/JiZCLN18}. Results show that our approach achieves better performance.
Furthermore, the ablation study shows that the sequential cross-modal semantic graph with the multi-modal adjacency matrix can effectively facilitate MABSC. 

In summary, our main contributions are as follows: 
We propose a sequential cross-modal semantic graph construction method, which can crossly utilize fine-grained information from images and text. 
Besides, we propose an encoder-decoder method with a prompt template that could effectively utilize the sequential cross-modal semantic graph considering the target.
We perform comprehensive experiments and extensive analysis on two datasets illustrating that SeqCSG can effectively and robustly model the multi-modal representations of descriptive texts and images and achieves state-of-the-art performance. 


\vspace{-10pt}
\section{Related Work} 
\zw{\textbf{Text-based Target-oriented Sentiment Classification.}}
This task aims to predict the sentiment polarities of the target, which is a mentioned entity in the text.
Dai et al.\cite{DBLP:conf/naacl/DaiYSLQ21} leveraged RoBERTa to reconstruct dependency trees, Yan et al. \cite{DBLP:conf/acl/YanDJQ020} proposed a generative framework that achieves competitive performance.

\zw{\noindent \textbf{Multi-modal Sentiment Classification}}
The goal of this task is to discover the sentiment expressed in multi-modal samples.  
Yu et al.\cite{DBLP:conf/aaai/YuXYW21} proposed the task of multi-modal joint training and learning multi-modal and unimodal representation
; 
 Yang et al.\cite{DBLP:conf/mm/YangXG20} extended the BERT model to cross-modal scenarios and proposed a multi-modal BERT for sentiment analysis; Wu et al.\cite{DBLP:journals/kbs/WuPZZTYMH22} designed a multi-modal emotion analysis model based on multi-head attention
 ; Keswani et al.\cite{DBLP:conf/semeval/KeswaniSAM20} 
 used BERT's multi-modal Bitransformer and ResNet to model text and visual features.
There were also some existing works that used LXMERT and ViLT \cite{DBLP:conf/icml/KimSK21} as the backbone for multi-modal sentiment analysis.

\zw{\noindent \textbf{Multi-modal Aspect-based Sentiment Classification}}
Xu et al.\cite{DBLP:conf/aaai/XuMC19} and Yu at el.\cite{DBLP:journals/taslp/YuJX20} used LSTM to effectively model the target-text and target-image interactions.
\cite{DBLP:conf/ijcai/Yu019,DBLP:conf/prcv/WangLSSQ21,DBLP:journals/www/ZhangWLLGY21,DBLP:conf/mm/0001F21} explored the usefulness of the BERT and proposed TomBERT, SaliencyBERT, ModalNet-BERT and EF-CapTrBERT.
Yu et al.\cite{DBLP:conf/ijcai/YuWXL22} proposed a multi-task learning model 
to leverage two auxiliary tasks
 to capture the image-target matching relations.
Zhao et al.\cite{DBLP:conf/coling/ZhaoWLDHC22} leveraged the adjective-noun pairs to align text and image.
The work most related to ours is VLP-MABSA \cite{DBLP:conf/acl/LingYX22}, which is a task-speciﬁc vision-language pre-training framework. 

\vspace{-10pt}
\section{Methodology}
Given a target entity mention $t$, a sentence $s$ where $t$ is located, and an image $v$ which is associated with $s$, MABSC aims to predict the sentiment label $y$ for $t$, where $ {y} \in \{ negative, neutral, positive\}$. $s$ is composed of a sequence of words, denoted as $\{{w}_1, {w}_2, {w}_3, \dots, {w}_{|N|}\}$, where $N$ is the sequence length, and $t$ can consist of multiple words. In the example shown in Figure \ref{fig:intro}.

Given a sample $m = \{ s, v, t, y\}$,  there are two steps in our method. 
First, we construct a sequential cross-modal semantic graph in order to represent the input of multi-modal information in the form of text. Our sequential cross-modal semantic graph elements consist of the tweet text, the caption, and the triples in the scene graph. For the input image ${v} \in \mathbb{R}^{3\times H \times W}$, we generate a caption of the image $v$, while a scene graph is extracted from the image $v$ via the scene graph generation method. Then, we input the semantic graph and multi-modal adjacency matrix into an encoder-decoder framework.
We introduce the graph construction in Sec. 3.1 and the encoder-decoder architecture in Sec. 3.2.


\begin{figure}[t]
\centering
\includegraphics[width=0.95\textwidth]{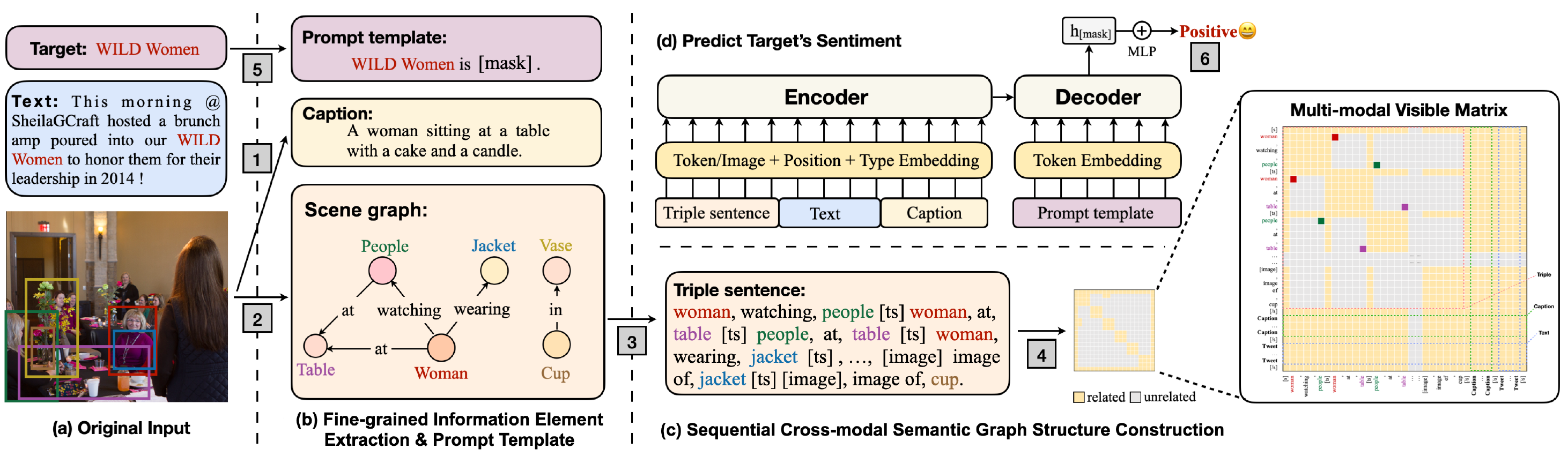}
\vspace{-5pt}
\caption{Overview of our proposed approach.
}
\vspace{-10pt}
\label{architecture}
\end{figure}


\subsection{Sequential Cross-modal Semantic Graph Construction}
The components of our sequential cross-modal semantic graph consist of scene graphs, image captions, and tweet text, which were carefully selected for their ability to provide a comprehensive representation of the visual content.
  
\noindent\textbf{Semantic Graph Element Extraction.}
There are three elements in our semantic graph: image caption, tweet text and scene graph.
We generate a caption of the image $v$ via caption transformers, the image captions serve to encapsulate global visual information while tweet text is already in the form of text.
In contrast to prior studies\cite{DBLP:conf/nips/LuBPL19} that rely on visual knowledge~\cite{Pan2017} sourced from object representations extracted from the image, 
we use scene graph, which consists of the Recall$@$5 $ (subject, predicate, object)$ triples from a pre-trained scene graph generator to represent the object-level image context, e.g., $ (car, behind, man)$, as well as Recall$@$5 $ (\left[ \tt{img} \right], image\  of, object)$ triples~\cite{Pan2009} to represent the relation between the sub-image and the object. Particularly, $\left[ \tt{img} \right] $ is a special token, which represents the relevant sub-image. The scene graphs were employed to depict local fine-grained image features.

\noindent\textbf{Element-to-Sequence Transformation.}
The merging of the caption, tweet text, and triple sentence is facilitated through the utilization of the separator token $\left[ \tt{/s} \right] $. This unified input is subsequently employed as the encoder input within the sequence-to-sequence model\cite{DBLP:journals/corr/abs-2207-12888}, conforming to the specified template:
\vspace{-4pt}
\begin{equation}
\small
    \zw{S_{in}} = \text{$\left[ \tt{s} \right] $} \ 
	\textbf{ triple sentence } \ \text{$\left[ \tt{/s} \right] $}  \ 
	\textbf{ caption } \text{$\left[ \tt{/s} \right] $} \ 
	\textbf{ tweet } \ 
	\text{$\left[ \tt{/s} \right] $} .
     \vspace{-1pt}
\end{equation}
The objective of our research is to establish a uniform sequence from the three elements in order to act as input to an encoder-decoder framework. 
Given that the tweet text and image caption are already presented in a sequential format, the primary objective of our transformation efforts centers on the integration of the scene graph into a textual format.
Specifically, two types of triples include object-to-object and object-to-image are extracted from the image, which are converted into serialized sentences separated by ``,'' and then connected via the special token $\left[ \tt{ts} \right]$ to construct those final triple sentences.
For example, given triples of the scene graph 
: $ (train$ ,$has$ ,$seat)$, $ (person$, $watching$, $man)$, $ (img_1$, $image~of$, $train)$, $ (img_2$, $image~of$, $person)$, $ (img_3$, $image~of$, $man)$. We convert them into the following serialized form:
\vspace{-2pt}
\begin{equation}
\small
	\text{$\left[ \tt{s} \right] $} \ \tt{ train, has, seat} \ \text{$\left[ \tt{ts} \right] $ ... }  \text{$\left[ \tt{ts} \right] $} \  \text{$\left[ \tt{img} \right] $} \tt{, image\ of, man }  \text{ $\left[ \tt{/s} \right] $} .
    \vspace{-2pt}
\end{equation}

\noindent\textbf{Semantic Graph Structure Construction.}
Our representation of a set of triples $\mathcal{T}_{in}$ entails the conversion of the set into a sequence of tokens. Despite the fact that the serialized triple sentence presently contains a significant amount of information concerning the triples, this serialization process is prone to damaging the inherent structure of the triple itself and compromising the implicit information that exists between entities.
Notably, one potential issue that arises with knowledge is the possibility that it may result in an alteration of the meaning conveyed within the original sentence.
Consequently, our objective is to ensure that the model enhances the internal connections present within the triples, while concurrently extracting additional valid information from the same entity within the serialized triple sentence.

Inspired by K-BERT \cite{DBLP:conf/aaai/LiuZ0WJD020}, we build a adjacency matrix to establish the interrelatedness between elements presented within the sequence of the semantic graph.
Formally, the adjacency matrix $M$ is defined as Eq. (3),
\vspace{-3pt}
\begin{equation}
\small
    M_{i j}=\left\{\begin{array}{l}
    1 \quad \text{ if } {w}_{i},{w}_{j} \in ({e}_{1}, {r}_{1}, {e}_{2}),  \\
    1 \quad \text{ if } {w}_{i} \in  K  \text{ or } {w}_{j} \in  K,  \\ 
    1 \quad \text{ if } ({w}_{i} \in S \cup C) \text{ or } ({w}_{j} \in S \cup C),\\ 
    1 \quad \text{ if } ({w}_{i} \in {e}_{1}) \cap ({w}_{j} \in {e}_{2}) \cap ({e}_{1}={e}_{2}), \\
    0 \quad \text { otherwise },
    \end{array}\right.
    \vspace{-3pt}
\end{equation}
where ${w}_{i}$ and ${w}_{j}$ are tokens in sentences; ${e}_{1}$ and ${e}_{2}$ are entities; ${r}_{1}$ is a relation; $K$ is special tokens; $S$ denotes the tweet text and $C$ denotes the image caption.

Concretely,
\textit{ (i)} for input triple sentences $S_{in}$, 
we make elements in the same triple visible to each other. The shared entities within various triples are visible to each other while the rest of is invisible. Through this approach, we mitigate the influence of extraneous information and effectively model implicit information present between entities;
\textit{ (ii)} the tweet, caption, and other special tokens in the encoder should be visible to each other so that the text information can interact with the triple information extracted from the image. 
To some degree, the adjacency matrix $M$ contains the structural information of the triple sentence.

\vspace{-10pt}
\subsection{Model Architecture}
\vspace{-5pt}
Our study employs a sequence-to-sequence architecture to implement a generative model, intended to classify the target's sentiment for MABSC. This approach is structured  two integral components, the encoder and the decoder. 
The overview of the model is shown in Figure \ref{architecture}.

\noindent\textbf{Encoder.}
The input of the encoder is composed of the sequential cross-modal semantic graph, which consists of three elements: scene graph, caption, and tweet text. For sentence $S_{in}$ in the encoder, we tokenize it into a sequence of tokens $S_{in} = \{s_1, s_2, ..., s_n\}$. 
The encoder is to encode sentence $S_{in}$ and adjacency matrix $M_{ij}$ into the hidden representation space as a vector $H_{en}$,
\vspace{-5pt}
\begin{equation}
\small
     H_{en} = Encoder (S_{in}, \zw{M_{ij}}), 
     \vspace{-5pt}
\end{equation}
where $H_{en} \in \mathbb{R}^{n\chenzhuo{\times}d}$ and $d$ is the hidden state dimension. 

To utilize the adjacency matrix, we make the encoder transformer layer aware of the relatedness between elements defined in $M$ in the self-attention module. 
The vanilla transformer layer includes a self-attention module and a position-wise feed-forward network. Suppose the input of self-attention module is $H=[ s_1, ..., s_n]^\top \in \mathbb{R}^{n\times d}$ with the ${i^{th}}$ row as the $d$ dimensional hidden state for the ${i^{th}}$ element. 
The self-attention operation is 
\vspace{-5pt}
\begin{equation}
\small
\begin{aligned}
Q &=H W_{Q}, K=H W_{K}, V=H W_{V}, 
\end{aligned}
\end{equation}
\vspace{-10pt}
\begin{equation}
\small
\begin{aligned}
A &=\frac{Q K^{\top}}{\sqrt{d_{K}}}, Attn (H)=softmax (A) V, 
\end{aligned}
\end{equation}
where $W_{Q}\in \mathbb{R}^{d\times d_Q}$,$W_{K}\in \mathbb{R}^{d\times d_K}$,$W_{V}\in \mathbb{R}^{d\times d_V}$ is the projection matrix to generate the query, key, and value representation of $H$ respectively; $A$ is the matrix capturing similarity between the query and the key. To inject adjacency matrix, we modify the self-attention module into
 \vspace{-5pt}
\begin{equation}
\small
A_{i j}=\frac{M_{i j} \times\left (h_{i} W_{Q}\right)\left (h_{j} W_{K}\right)^{\top}}{\sqrt{d}}+\left (1-M_{i j}\right) \time \delta,
\end{equation}
 \vspace{-10pt}
\begin{equation}
\small
{Attn}\left (h_{i}\right)=\sum\nolimits_{j=1}^{n_{s}} softmax\left (A_{i}\right)_{j} \times \left (e_{j} W_{V}\right),
 \vspace{-3pt}
\end{equation}
where $\delta$ is a large negative number to make values after the softmax function $softmax ()$ near $0$. 

The primary purpose of the embedding layer is to transform the sentence into an embedding representation that can be subsequently fed into the Transformers. Our proposed model adopts an approach similar to that of BERT\cite{DBLP:conf/naacl/DevlinCLT19}, wherein the embedding representation is calculated as a sum of three distinct embeddings, namely the element embedding, position embedding, and type embedding.

Our model contains language tokens and sub-image tokens. Therefore, the embedding process for our input is crucial in order to preserve its structural information.
Considering the input of multi-modal information, token/image embedding distinguishes input tokens. For text tokens, the vocabulary provided by BART \cite{DBLP:conf/acl/LewisLGGMLSZ20} is adopted. 
Each token in the sentence tree is mapped to an embedding vector with a dimension of H through a trainable lookup table. In addition, image tokens are encoded using ResNet and transformed into an embedding vector of the same dimension through a linear layer.

Following ViLT \cite{DBLP:conf/icml/KimSK21}, we set the image token embedding as 1 and the text token embedding as 0. In the context of transformer models, the absence of position embedding can cause the loss of structural information, leading to a bag-of-words model with unordered tokens. To avoid this issue, we adopted the position embedding technique used in the BART model for encoding purposes.

\noindent\textbf{Decoder.}
At the $t$-th time of decoding, the decoder takes the encoder's output $H_{en}$ and the decoder's previous output $y_1$, $y_2, ...$, $y_{t-1}$ as inputs. Then the decoder outputs $y_t$, where $i$ in $ y_i$ indicates the token index. 
Existing studies \cite{DBLP:journals/corr/abs-2107-13586} have shown that answer engineering has a strong influence on the performance of prompt-tuning. 
The basis for classification in the MABSC is not solely reliant on textual and visual inputs, but also on the target being evaluated.
For example, given the tweet text ``Congrats to Men's 2016 Finish Line 500 Festival 5K winner, Jackson Swisher of Danville! His time-16:59.'' and its corresponding image, the sentiment tendency of ``Jackson Swisher'' is ``Positive'' but  ``Danville'' is ``Neutral''. 
Therefore, it is crucial to consider the target during the integration and fusion of text and image information. To this end, we 
\zw{propose}
transforming the target information in the input into a prompt template. This approach enables the establishment of a connection between the target and sentiment orientation, resulting in a more accurate classification outcome.

Taking ``Congrats to Men's 2016 Finish Line 500 Festival 5K winner, Jackson Swisher of Danville! His time-16:59.'' as an example, the input content of the encoder remains the same and is composed of three elements: serialized triple sentence, caption, and tweet text. We transform the target ``Jackson Swisher'' in this sentence into the form of ``Jackson Swisher is $\left[ \tt{mask} \right] $ .'', input it to the decoder end, 
and then input the vector $H_{[m]}$ corresponding to the $\left[ \tt{mask} \right] $ in the last layer of the decoder into a MLP for sentiment classification.

Then for each target $x_{in}$ and the prompt template  $\mathcal{T}$, let the manipulation $X_{prompt} = \mathcal{T} (x_{in})$ be a masked language modeling (MLM) input which contains one $\left[ \tt{mask} \right] $ token. In this way, we can treat our task as a MLM, and model the probability of predicting class $y \in \mathcal{Y}$ as:
 \vspace{-3pt}
\begin{equation}
\small
p\left (y \mid H_{\left[ m \right]}\right)={softmax}\left (\theta_{Linear} {Dropout}\left (H_{\left[ m  \right]}\right)\right), 
 \vspace{-3pt}
\end{equation}
where $H_{\left[ m \right]}$ is the hidden vector of $\left[ \tt{mask} \right] $. 
$\theta_{Linear} \in \mathbb{R}^{3 \times 768}$ is learned by back propagation. We learn $\theta_{Linear}$ by fine-tuning the BART alongside Eq.(9) using the standard cross-entropy loss.

\vspace{-10pt}
\section{Experiments}
\vspace{-10pt}
In this section, we compared with one image-only, five text-only and several text-image baselines to demonstrate the effectiveness of our method by answering the following questions:
\noindent\textbf{Q1: } How does SeqCSG perform compared with state-of-the-art methods for MABSC?
\noindent\textbf{Q2: } Do image captions and scene graphs help capture the fine-grained information of images better?
\noindent\textbf{Q3: } Whether the multi-modal adjacency matrix help crossly utilize image and text information?


\vspace{-7pt}
\subsection{Experiment Setting}
\vspace{-7pt}
\noindent\textbf{Datasets.} 
\zw{We conduct experiments on two benchmarks Twitter2015 and Twitter2017 \cite{DBLP:conf/ijcai/Yu019}.}

\noindent\textbf{Implement Details.}
We employ BART \cite{DBLP:conf/acl/LewisLGGMLSZ20}, a denoising and simple encoder-decoder PLM. 
The image caption is obtained by the transformer-based caption model \cite{DBLP:conf/mm/0001F21}. 
We utilize a pre-trained scene graph generator \cite{DBLP:conf/cvpr/TangNHSZ20} to extract a scene graph.
Note that we freeze the ResNet parameters to decrease the learnable parameters hence avoiding overfitting. 
Specifically, we ﬁx all the hyper-parameters after tuning them on the development set and ﬁne-tune for 30 epochs. The batch size is set to 16; the learning rate is set to 2e-5. We implement all the models with PyTorch, and run experiments on a RTX3090 GPU.

\vspace{-10pt}
\begin{table*}[t]
    \caption{The property prediction performance of our method (SeqCSG), compared with image-only (first group), text-only  (second group) and multi-modal methods (third group)  baselines on Twitter2015 and Twitter2017 datasets.
    }
    \setlength\tabcolsep{12pt}
    \renewcommand\arraystretch{0.7}
    \centering
    \resizebox{\linewidth}{!}{
    \begin{tabular}{c|lcccc}
        \toprule
        \multirow{2}{*}{\textbf{Modality}} & \multirow{2}{*}{\textbf{Method}} & \multicolumn{2}{c}{\textbf{Twitter2015}} & \multicolumn{2}{c}{\textbf{Twitter2017}} \\
        \cline{3-4} \cline{5-6}
        & & Acc & Macro-F1 & Acc & Macro-F1 \\
        \hline
        \textbf{Visual} & Res-Target & 59.9 & 46.5 & 58.6 & 54.0 \\

        \hline
        \multirow{4}{*}{\textbf{Text}} 
        & MGAN \cite{DBLP:conf/emnlp/FanFZ18} & 71.2 & 64.2 & 64.8 & 61.5 \\
        & BERT \cite{DBLP:conf/naacl/DevlinCLT19} & 74.3 & 70.0 & 68.9 & 66.1 \\
        & BERT+BL \cite{DBLP:conf/naacl/DevlinCLT19} & 74.3 & 70.0 & 68.9 & 66.1 \\
        & BERT-Pair-QA \cite{DBLP:conf/naacl/SunHQ19} & 74.4 & 67.7 & 63.1 & 59.7 \\
        & BART \cite{DBLP:conf/acl/LewisLGGMLSZ20} & 76.0 & 67.6 & 69.5 & 67.0 \\
        
        \hline
        \multirow{18}{*}{\textbf{Text + Visual}} 
        & Res-MGAN & 71.7 & 63.9 & 66.4 & 63.0 \\
        & Res-BERT+BL & 75.0 & 69.2 & 69.2 & 66.5 \\
        & mPBERT (CLS) \cite{DBLP:conf/ijcai/Yu019} & 75.8 & 71.1 & 68.8 & 67.1 \\
        & TomBERT \cite{DBLP:conf/ijcai/Yu019} & 77.2 & 71.8 & 70.5 & 68.0 \\
        & MIMN \cite{DBLP:conf/aaai/XuMC19} & 71.8 & 65.7 & 65.9 & 63.0\\
        & ViLBERT \cite{DBLP:conf/nips/LuBPL19} & 73.8 & 69.9 & 67.4 & 64.9 \\
        & ModalNet-BERT \cite{DBLP:journals/www/ZhangWLLGY21} & 79.0 & 72.5 & 72.4 & 69.2 \\
        & CapTrBERT \cite{DBLP:conf/mm/0001F21} & 78.0 & 73.2 & 72.3 & 70.2 \\
        & JML-MASC \cite{DBLP:conf/emnlp/JuZXLLZZ21} & 78.7 & - & 72.7 & - \\
        & SaliencyBERT \cite{DBLP:conf/prcv/WangLSSQ21} & 77.0 & 72.4 & 69.7 & 67.2 \\
        & VLP-MABSA \cite{DBLP:conf/acl/LingYX22} & 78.6 & 73.8 & 73.8 & 71.8 \\
        & ITM \cite{DBLP:conf/ijcai/YuWXL22} & 78.3 & 74.2 & 72.6 & 72.0 \\
        & KEF-SaliencyBERT \cite{DBLP:conf/coling/ZhaoWLDHC22} & 78.2 & 73.5 & 71.9 & 69.0 \\
        & KEF-TomBERT \cite{DBLP:conf/coling/ZhaoWLDHC22} & 78.7 & 73.8 & 72.1 & 70.0 \\
        \cline{2-6}
        & Multi-BART \cite{DBLP:conf/acl/LewisLGGMLSZ20} & 77.2 & 72.6 & 70.5 & 69.0 \\
        & \textbf{SeqCSG (Ours)} & \textbf{79.3} & \textbf{75.0} & \textbf{74.6} & \textbf{73.2} \\
        
        \bottomrule
    \end{tabular}
    }
    \label{tab: sentiment classification result}
    \vspace{-10pt}
\end{table*}

\subsection{Main Results (Q1)}
\vspace{-3pt}
Table \ref{tab: sentiment classification result} shows the results of different methods on both Twitter2015 and Twitter2017.
Based on the results of our experiments, SeqCSG has demonstrated superior performance compared to other baseline models across all benchmark datasets. 
Notably, our model achieves a greater F1-score by 1.2 and 1.4 absolute percentage points, respectively, and accuracy that is 0.7 and 0.8 absolute percentage points higher, respectively, than the VLP-MABSA system.

Our approach stands out due to the utilization of image captions and scene graphs, which allow for the modeling of both global and local fine-grained information present in the original image. By processing these elements through a multi-modal adjacency matrix alongside the tweet text, we are able to extract a significant amount of auxiliary information from the image, including the relationship between entities and relevant sub-images. This enables our model to learn an implicit correlation representation of the target, fine-grained information, and tweet text during training, which leads to superior performance compared to other methods.
In our approach, we aim to mitigate the negative impact of triple knowledge noise. To achieve this, we set a limit on the number of triples, while also taking care to ensure that the serialized triple knowledge only interacts with each other when a connection exists through the multi-modal adjacency matrix. This analysis indicates \textbf{SeqCSG performs good for MABSC compared with other methods (Q1)}. 

We observe that the performance of single-modal methods, either based on image or text alone, is inferior to that of their multi-modal counterparts. Specifically, the image-based methods exhibit much lower accuracy than the multi-modal approaches, while the text-based methods also suffer from a certain performance gap. Our findings suggest that there is still considerable potential for improving the processing of visual features as well as enhancing the interaction between modalities to achieve better results.

From the results, we can observe that Multi-BART also achieves a good performance, even better than some multi-modal methods. This observation serves as evidence of the effectiveness of the proposed framework as a solid foundation.
In the context of multi-modal methods, VLP-MABSA outperforms its predecessors due to its design of pre-training tasks tailored to specific tasks, thereby facilitating alignment and interaction between textual and visual features.



\vspace{-10pt}
\subsection{Ablation Study (Q2 \& Q3)}
\vspace{-3pt}
\noindent\textbf{Component Analysis.}
We perform an ablation study to evaluate the efficacy of each component on Twitter2015.
Results are shown in Table \ref{tab:Ablation}.
\vspace{-10pt}
\begin{table}[htbp]
    \caption{Ablation Study on Twitter2015 dataset. 
    }
    \centering
    \setlength\tabcolsep{12pt}
    \renewcommand\arraystretch{1}
    \resizebox{0.6\linewidth}{!}{
    \begin{tabular}{lcc}
    \toprule
    {\bf Method}  & {\bf Acc}  & {\bf Macro-F1}  \\
    \toprule
    Multi-BART (CLS) & 77.2 & 72.6 \\
    w/o \{caption\} & 76.0 & 67.6 \\
    \hline
    \textbf{SeqCSG (Ours)} & \textbf{79.3} & \textbf{75.0}\\
    w/o \{caption\} & 77.0 & 72.8 \\
    w/o \{adjacency matrix\} & 78.9 & 74.7 \\
    w/o \{adjacency matrix \& scene graph\} & 78.2 & 74.4\\
    w/o \{prompt\} & 78.4 & 74.2 \\
    w/o \{freeze\} & 78.6 & 74.3 \\    
    \bottomrule
    \end{tabular}
    }
    \label{tab:Ablation}
    \vspace{-10pt}
\end{table}


An important distinction between BART and Multi-BART lies in the latter's inclusion of image caption information as input. Upon removing the caption and using only triple sentences and text, we observe a decrease of [2.3, 2.2] points in [acc, F1] performance. These comparative results suggest that the image caption serves as a valuable global representation.
As an essential component, it is evident that the experimental results show a significant decrease without the presence of scene graphs, as compared to the SeqCSG model.
All these observations verify that \textbf{both image captions and scene graphs help capture fine-grained information of image better
(Q2).}

Our results prove that the incorporation of the multi-modal adjacency matrix enhances the performance, indicating the efficacy of serializing the sequential cross-modal semantic graph in conjunction with the adjacency matrix. Therefore we can draw the conclusion that the \textbf{multi-modal adjacency matrix can make crossly utilize image and text information usefully. (Q3)}

Instead, SeqCSG optimizes the input structure of the model and converts the problem into a classification problem under a generation-based paradigm. We observe that our model exhibits a performance decay in the absence of other components, i.e., prompt template, parameter frozen, indicating the efficacy of all the modules.
Concretely, we observe that taking the sequence-to-sequence model as the base comparison, our model achieves significant improvement (4.3\% on accuracy and 10.9\% on f1-score), which verifies its effectiveness. The performance rises sharply when taking the prompt template built for the target as the input to the decoder side of the model. We argue that the design of prompt templates, along with aspect-based sentiment classification using embeddings corresponding to the $\left[ \tt{mask} \right] $ , highly appropriate for this specific scenario. By utilizing the prompt template, we are able to establish an implicit association between the multi-modal corpus and the target, thereby enabling targeted classification predictions for multiple targets in a sentence.
\begin{figure}[htbp]
  \centering
  \includegraphics[width = 0.75\linewidth]{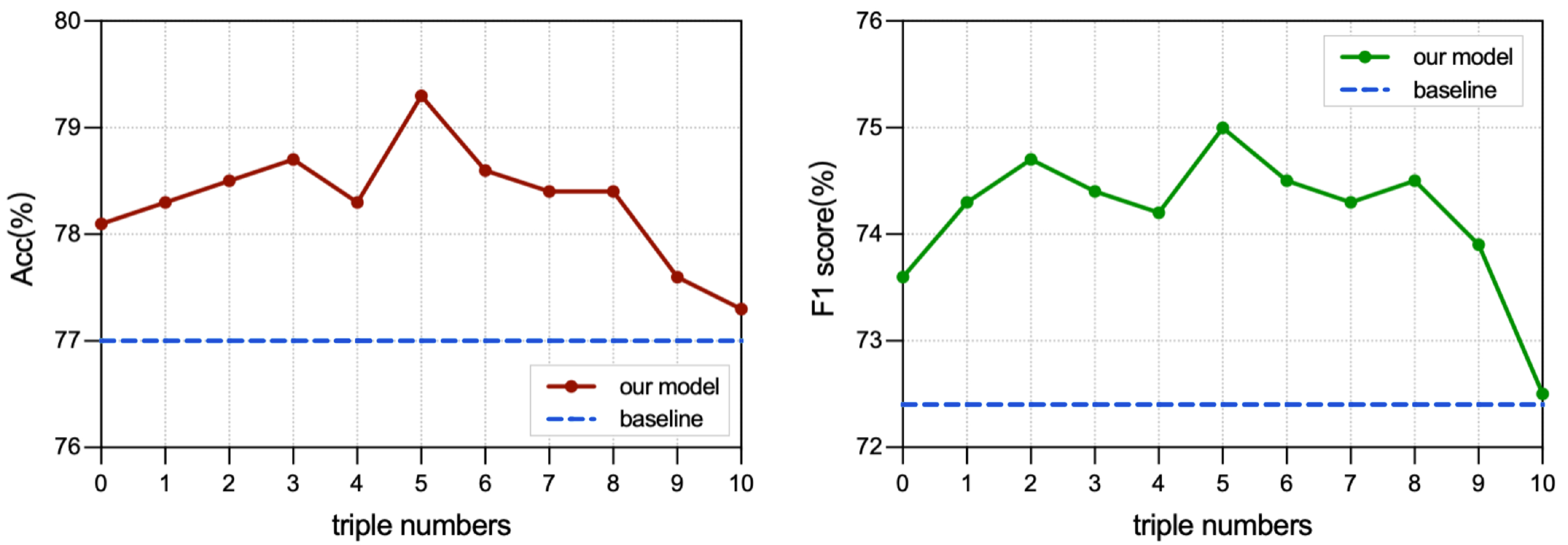}
    \vspace{-5pt}
  \caption{Performance of SeqCSG with different triple numbers on Twitter2015.}
  \label{fig:ablation2}
\end{figure}

\noindent\textbf{Impact of the triple numbers.}
Upon conducting an analysis of the triple numbers with Figure \ref{fig:ablation2}. We draw the following conclusions: Feeding the triples into the model has a certain performance gain. The performance of SeqCSG is highly influenced by both the quantity and quality of triples. The model performs best when the number of recalled triples is controlled to five triples 
A lower number of triples may limit the availability of detailed image information contained in the triples. Conversely, an excessively high number of triples may impede the efficiency of the training process and increase the risk of noisy data.

\vspace{-10pt}
\begin{figure*}
\centering
\includegraphics[width=0.95\textwidth]{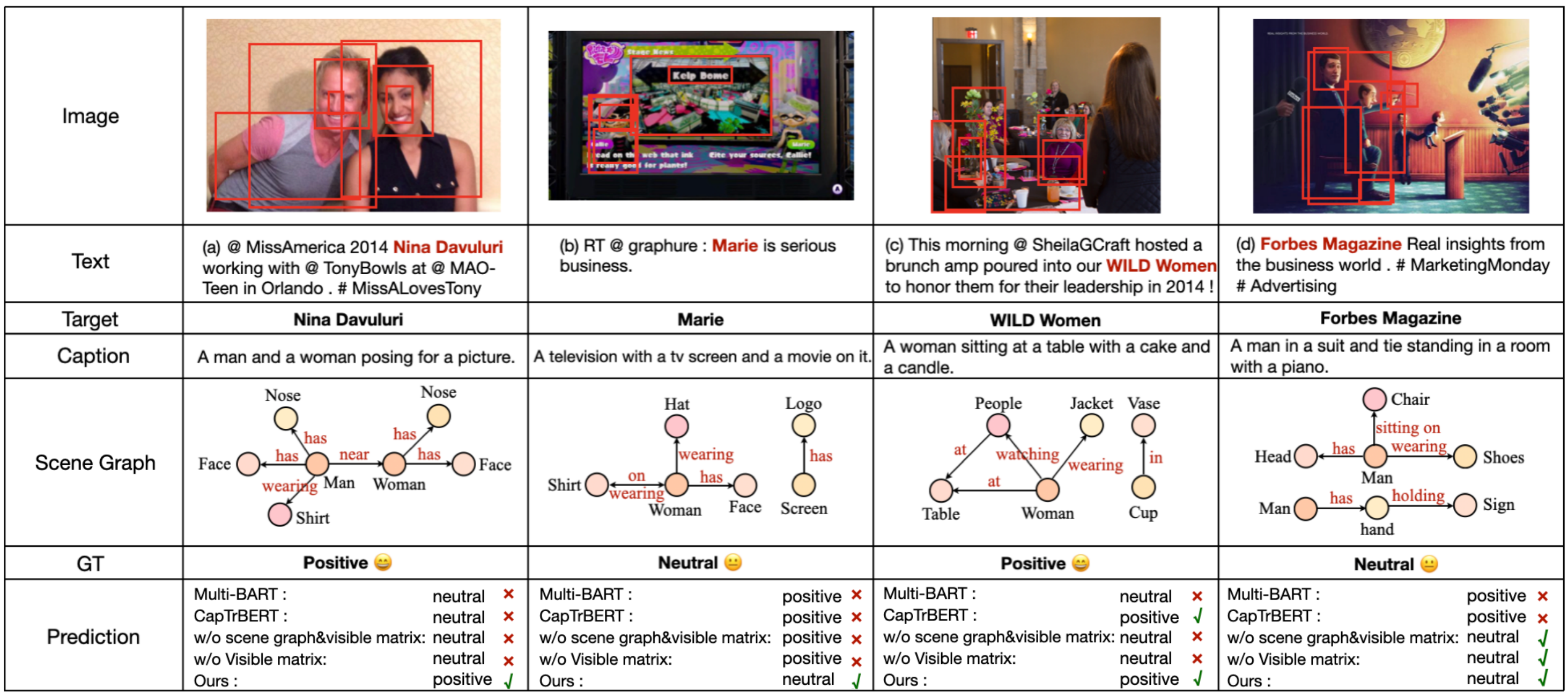}
\vspace{-5pt}
\caption{Predictions of different approaches.
}
\label{fig:case_study}
\vspace{-10pt}
\end{figure*}

\vspace{-10pt}
\subsection{Case Study}
\vspace{-5pt}
To further analyze the robustness of SeqCSG for error sensitivity, we visualize some predictions from different methods. 
The compared methods include BART, CapTrBERT, our model using the same inputs without scene graph and adjacency matrix, and our model using scene graph without adjacency matrix, respectively.
As illustrated in Figure\ref{fig:case_study}, BART outputs wrong predictions in all these four cases. CapTrBERT outputs correct prediction in the third case but makes mistakes in the first, second and fourth cases,  where the caption can not provide enough information from images. In contrast, our full model, which combines the scene graph and the adjacency matrix, makes correct predictions in those cases.
Among all the cases, we notice that SeqCSG obtains more fine-grained image representation, which is essential for reducing error sensitivity. We can further reveal that the model lacking a scene graph and adjacency matrix has a poor prediction effec, which shows the superiority of our framework and the multi-modal adjacency matrix crossly utilizes image and text information.

\vspace{-10pt}
\section{Conclusion}
\vspace{-5pt}
In this paper, we propose a multi-modal aspect-based sentiment classification (MABSC) method SeqCSG where the sequential cross-modal semantic graphs are constructed to support our encoder-decoder sentiment classification framework. 
Experimental results show that our proposed approach generally outperforms the state-of-the-art methods on standard benchmarks. As a unified model, SeqCSG integrates the advantages of prompts and sequential cross-modal semantic graphs to effectively model global and local fine-grained image information and crossly utilize image and text information.
\vspace{-10pt}



%
%

\bibliographystyle{splncs04}
\bibliography{references}

\newpage
\appendix
\section*{Appendix}

\section{Details of SeqCSG}

\subsection{Datasets}
We adopt two datasets from \cite{DBLP:conf/ijcai/Yu019}, namely Twitter2015 and Twitter2017, for evaluation. 
\zw{We conduct experiments on two benchmarks Twitter2015 and Twitter2017 \cite{DBLP:conf/ijcai/Yu019}.}
Their statistics are shown in Table \ref{tab:data_statistics}. 
Both datasets contain multi-modal tweets, each of which consists of a text, an image posted alongside the tweet text, annotated targets within the tweet text, and annotated sentiments of the targets. Each sentiment is a label from $\{negative, neutral, positive\}$.
\setlength{\tabcolsep}{10pt}
\begin{table}[htbp]   
\caption{The basic statistics of two Twitter datasets. 
} 
\label{tab:data_statistics}
\scriptsize 
\centering
\resizebox{0.9\linewidth}{!}{
\begin{tabular}{lcccccc}
    \toprule
    &\multicolumn{3}{c}{Twitter2015}&\multicolumn{3}{c}{Twitter2017} \\
    \cline{2-7}
    & \textbf{Train} & \textbf{Dev} & \textbf{Test}  & \textbf{Train} & \textbf{Dev} & \textbf{Test} \\ 
    \hline
    Positive     &928	    &303	&317	&1508	&515	&493 \\ 
    Neutral      &1883	&670	&607	&1638	&517	&573 \\
    Negative     &368	    &149	&113	&416	&144	&168 \\  
    \midrule
    Total Aspects      &3179	&1122	&1037	&3562	&1176	&1234 \\ 
    \midrule
    \text{\#Sentence} &2101 &727 &674 &1746 &577 &587\\
    \text{\#Targets} &1.3 &1.3 &1.3 &1.4 &1.4 &1.4\\
    \text{\#Length} &16.7 &16.7 &17 &16.2 &16.4 &16.4 \\
    \bottomrule
    \vspace{-10pt}
\end{tabular}
}
\end{table}

\subsection{Data Pre-processing}
The input of the encoder consists of the triple sentence, caption and tweet text. The caption is mainly to capture the global information of the image. Considering that the tweet text comes from Twitter, the text contains many special symbols, we perform data cleaning on it. Meanwhile, we replace the 
\$$T$\$ representing the target in the text with the specific target name, and add $\left[ \tt{target} \right] $ and $\left[ \tt{/target} \right] $ before and after the specific target in the text. The purpose is to guide the model to focus on the target and target-related content.
The triple sentence consists of serialized triples separated by a special token $\left[ \tt{ts} \right] $. And each triple is converted into text by connecting the head entity, relationship, and tail entity with a comma.

Considering that the above three parts together constitute the input of the encoder, we connect them in the following two ways,
\begin{equation}
\small
\begin{aligned}
    & \text{$\left[ \tt{s} \right] $} \ 
	\text{$\left[ \tt{triple} \right] $} \ \textbf{ triple sentence } \ \text{$\left[ \tt{/triple} \right] $}  \ 
	\text{$\left[ \tt{caption} \right] $} \textbf{ caption } \\ & \text{$\left[ \tt{/caption} \right] $} \ 
	\text{$\left[ \tt{tweet} \right] $} \textbf{ tweet } \text{$\left[ \tt{/tweet} \right] $} \ 
	\text{$\left[ \tt{/s} \right] $}
\end{aligned}
\end{equation}
\begin{equation}
\small
\begin{aligned}
    \text{$\left[ \tt{s} \right] $} \ 
	\textbf{ triple sentence } \ \text{$\left[ \tt{/s} \right] $}  \ 
	\textbf{ caption } \text{$\left[ \tt{/s} \right] $} \ 
	\textbf{ tweet } \ 
	\text{$\left[ \tt{/s} \right] $}
\end{aligned}
\end{equation}

\subsection{Multi-modal Visible Matrix Construction}
In this section, we provide the details of the multi-modal visible matrix construction. We construct a relatedness matrix to indicate the relationship between relevant tokens for the input that contains serialized triples, tweets and captions on the encoder side.  

\begin{figure}[htbp]
  \centering
  \includegraphics[width = 0.95\linewidth]{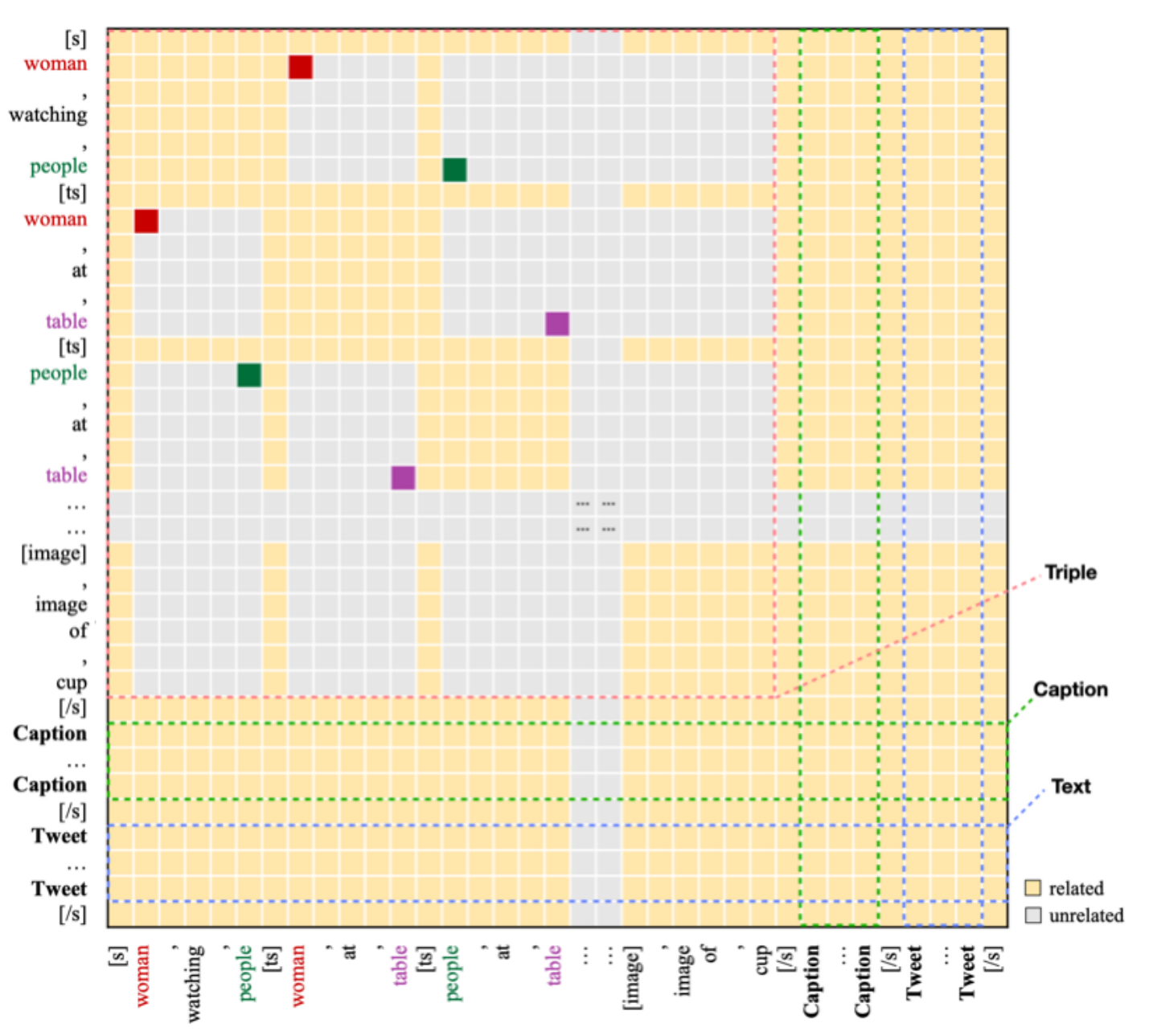}
  \caption{Visible matrix construction example to indicate the relationship between relevant tokens for the input that contains serialized triples, tweets and captions.}
  \label{fig:visible matrix}
  \vspace{-10pt}
\end{figure}

We formulate different rule constraints for serialized triple sentences, captions, tweet texts and special tokens. For the tweet text, caption and other special tokens, we require them to be visible, so that the text information can interact with the triple information generated by the image. The visible matrix between triple sentences is very critical because it can establish implicit relationships between targets while restoring the triple structure. In addition, restricting the invisible relationship between some entities and relationship tokens can reduce the noise of the model. The specific multi-modal visible matrix construction is shown in Figure \ref{fig:visible matrix}.

\section{Details about Experimental Setup}
\subsection{Dataset Descriptions}
We adopt two benchmark datasets annotated by \cite{DBLP:conf/ijcai/Yu019}, namely Twitter2015 and Twitter2017 for MABSC to evaluate our model.
Twitter2015 and Twitter2017 are two multi-modal datasets respectively collected by \cite{DBLP:conf/aaai/0001FLH18} and \cite{DBLP:conf/acl/JiZCLN18}.  
Both datasets are constructed similarly. Take Twitter2015 as an example, they use Twitter’s API to collect the tweets. The collection includes 26.5 million tweets. Then they drop the non-English tweets and extract containing relevant images from all those tweets, leaving 4.3 million tweets. They randomly sampled 50,000 data containing images from tweets covering various topics to reduce user-introduced specificity. Following the standard annotation naturally, annotators annotate entities whose entity types are \textbf{Person}, \textbf{Location}, \textbf{Organization}, or \textbf{Miscellaneous}. On this basis, in order to solve the MABSC task, \cite{DBLP:conf/ijcai/Yu019} asks three domain experts to annotate the sentiment towards each target, and take the majority label among the three annotators as the gold label. Basic statistics have shown in Table 3.

\subsection{Baselines}
We adopt three types of baselines. The details of each baseline are listed below:
\paragraph{Image Only Methods.}
\begin{itemize}
    \item \textit{Res-Target} ~ \cite{DBLP:conf/cvpr/HeZRS16} directly applies cross-modal attention to ResNet input features as the language features without any extra modifications.
\end{itemize}
\paragraph{Text Only Methods.}
\begin{itemize}
    \item \textit{MGAN} ~ \cite{DBLP:conf/emnlp/FanFZ18} proposes a fine-grained attention mechanism, which is responsible for linking and fusing the words from the aspect and context. Then this model combines it with the coarse-grained attention mechanism in order to capture the word-level interaction. 
    \item \textit{BERT} ~ \cite{DBLP:conf/naacl/DevlinCLT19} is a simple baseline that only uses BERT encoder.
    \item \textit{BERT+BL} ~ \cite{DBLP:conf/naacl/DevlinCLT19} is BERT with another BERT layer stacked on it.
    \item \textit{BERT-Pair-QA} ~ \cite{DBLP:conf/naacl/SunHQ19} uses the auxiliary question method to obtain SOTA on SemEval 2014 Task 4.
    \item \textit{BART} ~ \cite{DBLP:conf/acl/LewisLGGMLSZ20} is a baseline under a generation-based paradigm, which only takes text and target as input.
\end{itemize}
\paragraph{Text and Image Methods.}
\begin{itemize}
    \item \textit{Res-MGAN} ~ \cite{DBLP:conf/emnlp/FanFZ18} uses a multi-grain attention network for aspect understanding.
    \item \textit{Res-BERT+BL} ~ \cite{DBLP:conf/naacl/DevlinCLT19} directly applies cross-modal attention to ResNet input features and the language features without any extra modifications.
    \item \textit{TomBERT} ~ \cite{DBLP:conf/ijcai/Yu019} is also a target-oriented multi-modal BERT model. TomBERT builds on top of the baseline BERT architecture by adding target-sensitive visual attention and more self-attention layers to capture cross-modal dynamics.
    \item \textit{ModalNet-BERT} ~ \cite{DBLP:journals/www/ZhangWLLGY21} leverages two memory networks for mining the intra-modality information of text and image, and then design a discriminant matrix to supervise the fusion of inter-modality information.
    \item \textit{CapTrBERT} ~ \cite{DBLP:conf/mm/0001F21} optimizes on TomBERT, but transforms image information into image caption, and then fuses the information of the two modalities.
    \item \textit{SaliencyBERT} ~ \cite{DBLP:conf/prcv/WangLSSQ21} proposes a recurrent attention network over the BERT architecture.
    \item \textit{JML-MASC} ~ \cite{DBLP:conf/emnlp/JuZXLLZZ21} is a multi-task learning approach proposed recently with the auxiliary cross-modal relation detection task.
    \item \textit{VLP-MABSA} ~ \cite{DBLP:conf/acl/LingYX22} proposes a task-specific Vision Language Pre-training framework for MABSA. 
    \item \textit{ITM} ~ \cite{DBLP:conf/ijcai/YuWXL22} proposes a multi-task learning model named coarse-to-fine grained Image-Target Matching network, which leveraged two auxiliary tasks, i.e., ImageTarget Relevance and Object-Target Alignment, to capture the image-target matching relations.
    \item \textit{KEF} ~ \cite{DBLP:conf/coling/ZhaoWLDHC22} leverages the adjective-noun pairs to align text and image.
    \item \textit{Multi-BART} ~ \cite{DBLP:conf/acl/LewisLGGMLSZ20} is strong baseline under a generation-based paradigm. Unlike BART, Multi-BART uses the information of two modalities of image and text, and takes text, target and caption generated by image as the input of the model. 
\end{itemize}

\subsection{Downstream Details}
This section contains details about training procedures and hyper-parameters for each dataset. We utilize Pytorch to conduct experiments with a RTX3090 GPU. All optimizations are performed with the AdamW optimizer with a linear warmup of the learning rate. 

Specifically, We employ BART \cite{DBLP:conf/acl/LewisLGGMLSZ20}, a denoising and simple encoder-decoder PLM, to implement our method. The encoder and decoder both have six layers and are initialized with BART-base parameters. 
The image caption is obtained by the transformer-based caption model \cite{DBLP:conf/mm/0001F21}.
We utilize a pre-trained scene graph generator \cite{DBLP:conf/cvpr/TangNHSZ20} to extract a scene graph.
Note that we freeze the ResNet parameters to decrease the learnable parameters hence avoiding overfitting, leaving only one linear layer to learn. 
We detail the hyper-parameter as follows: 
\paragraph{Twitter2015}
\begin{itemize}
    \item max epoch: \textbf{30}
    \item batch size: [8, \textbf{16}]
    \item learning rate: [1e-5, \textbf{2e-5}, 3e-5]
    \item image encoder: [ResNet18, ResNet34, \textbf{ResNet50}, ResNet101]
    \item number of triples between objects: [0, 1, 2, 3, 4, \textbf{5}, 6, 7, 8, 9, 10]
    \item number of triples between object and image: [0, 1, 2, 3, 4, \textbf{5}, 6, 7, 8, 9, 10]
\end{itemize}

\paragraph{Twitter2017}
\begin{itemize}
    \item max epoch: \textbf{30}
    \item batch size: [8, \textbf{16}]
    \item learning rate: [\textbf{1e-5}, 2e-5, 3e-5]
    \item image encoder: [ResNet18, ResNet34, \textbf{ResNet50}, ResNet101]
    \item number of triples between objects: [0, 1, 2, 3, 4, \textbf{5}, 6, 7, 8, 9, 10]
    \item number of triples between object and image: [0, 1, 2, 3, 4, \textbf{5}, 6, 7, 8, 9, 10]
\end{itemize}
\section{Additional Experimental Results}
\subsection{Effect of Image Encoder}
We further analyze the influence of image encoders on Twitter2015. As shown in Table \ref{tab:image encoder}, We can draw the following conclusions: (1) Different image encoding modules have improved the effectiveness of the model. (2) 
Different ResNet encoders have different effects on the experimental results. This is mainly because the granularity of the sub-images we extract from the picture is different, which leads to different effective information brought by different encoders.

\begin{table}[htbp]
    \caption{Performance of different image encoders on Twitter2015 dataset for MABSC task.
    }
    \centering
    \setlength\tabcolsep{3pt}
    \renewcommand\arraystretch{1}
    \resizebox{0.6\linewidth}{!}{
    \begin{tabular}{lcc}
    \toprule
    {\bf Method}  & {\bf Acc}  & {\bf Macro-F1}  \\
    \toprule
	w/o \{scene graph \& visible matrix\} & 78.2 & 74.4\\
    \hline
    SeqCSG (ResNet18) & 78.7 & 75 \\
    SeqCSG (ResNet34) & 78.7 & 75.1 \\
	SeqCSG (ResNet50) & \textbf{79.3} & 75\\
	SeqCSG (ResNet101) & 78.9 & \textbf{75.4}\\
    \bottomrule
    \end{tabular}
    }
    \label{tab:image encoder}
    \vspace{-10pt}
\end{table}

\subsection{Interpretability Analysis}
\begin{figure}[htbp]
  \centering
  \includegraphics[width = 0.8\linewidth]{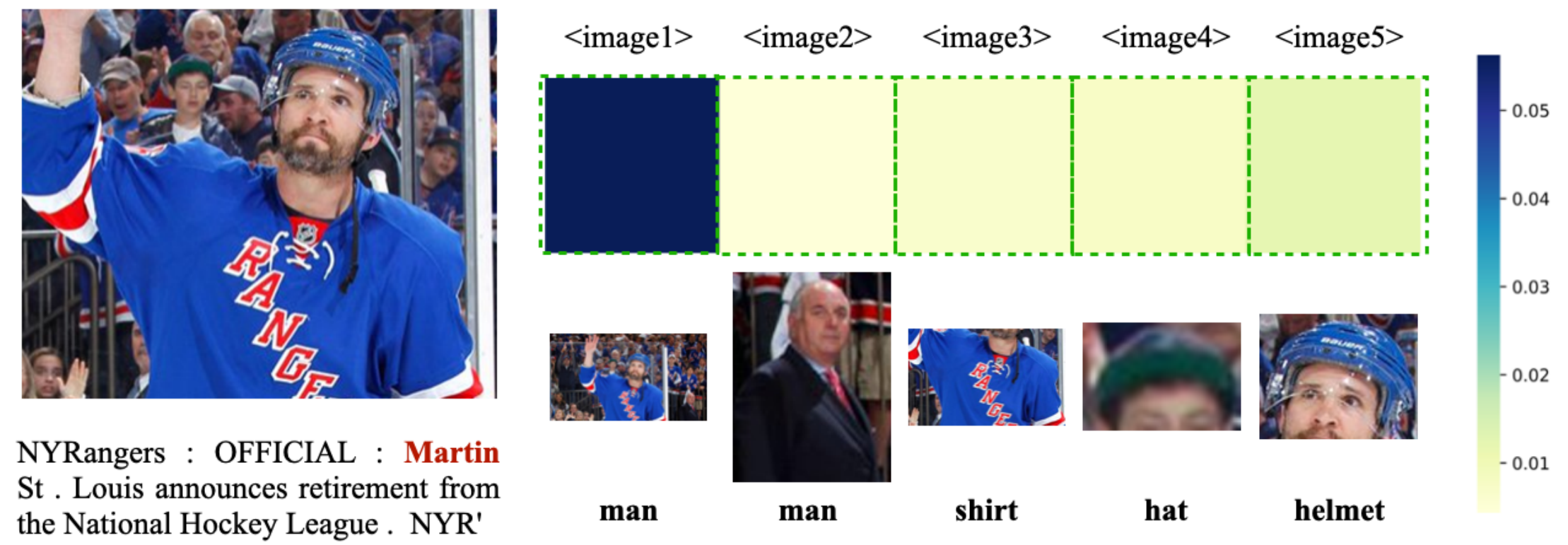}
  \caption{The corss attention visualizations between $\left[ img \right]$ token in encoder and $\left[ mask \right]$ token in decoder.}
  \label{fig:heatmap}
\end{figure}

Figure \ref{fig:heatmap} visualizes the cross attention between $\left[ img \right]$ token in the encoder and $\left[ mask \right]$ token in decoder maps on the case.
Through the visualization of the case, we can notice the cross attention weights reveal that our model can capture the fine-grained semantics of the image. More importantly, our model can learn the implicit correlation representation of the target and the relevant sub-image.
We can draw the conclusion that irrelevant visual features may hurt the performance, while our model is able to benefit from more fine-grained and implicit multi-modal representation, which is essential for reducing error sensitivity.

\section{Future Works}
In the future, we plan to 
\textit{ (i)} apply our approach to more image-enhanced natural language processing and information retrievals tasks, such as multi-modal event extraction and multi-modal entity retrieval;
\textit{ (ii)} incorporate the conceptual KG, and unify the image-level information and conceptual knowledge to perform joint reasoning of the scene, which is applied to tasks like image-text matching\cite{huang2023structure},visual question answering \cite{DBLP:conf/semweb/0007CGPYC21};
and \textit{ (iii)} put attention to low resource scenarios \cite{DBLP:conf/ijcai/ChenG0HPC21,DBLP:journals/corr/abs-2112-10006,DBLP:journals/corr/abs-2106-15047} with
less or even no training data.

\end{document}